\documentclass[letterpaper, 10 pt, conference]{ieeeconf}

\IEEEoverridecommandlockouts

\usepackage{graphics} \usepackage{graphicx}
\newif\ifanonymous
\anonymousfalse         \usepackage{booktabs}
\usepackage{color}
\usepackage{siunitx}
\usepackage{algorithm}
\usepackage{float}
\usepackage{amsmath}
\usepackage{amssymb}
\usepackage{xcolor}
\usepackage{cite}
\usepackage{balance}
\usepackage{longtable}
\usepackage{threeparttable}
\usepackage{multirow}
\definecolor{bshade}{rgb}{0.55,0.75,0.95}
\newcommand{\best}[1]{  \colorbox{bshade!100}{\makebox(12,3)[b]{\raisebox{-0.055pt}[0pt][0pt]{\textbf{#1}}}}}
\newcommand{\secondbest}[1]{  \colorbox{bshade!40}{\makebox(12,3)[b]{\raisebox{-0.055pt}[0pt][0pt]{#1}}}}
\title{\LARGE \bf
RiskFly: Frustum-Aligned Spatio-Temporal Risk Fields for \\ One-Stage Agile Flight in Dynamic Clutter
}
\ifanonymous
\author{}
\else

\author{Luxia Ai$^{1}$, Haopeng Chen$^{1}$, Yuchao Mei$^{1}$, Guohao Zhang$^{1}$, and Wenbing Tao$^{1}$ 
\thanks{*This work was not supported by any organization}
\thanks{$^{1}$ The authors are with the State Key Laboratory of Multispectral Information Intelligent Processing Technology, School of Artificial Intelligence and Automation, Huazhong University of Science and Technology, Wuhan 430074, China (e-mail: luxiaai@hust.edu.cn; m202473877@hust.edu.cn; m202473809@hust.edu.cn; m202573768@hust.edu.cn; wenbingtao@hust.edu.cn).}%
}

\fi
\renewcommand{\thesubsubsection}{\thesubsection\arabic{subsubsection}}
\begin{document}
\maketitle
\thispagestyle{empty}
\pagestyle{empty}

\begin{abstract}
Agile flight in unknown, cluttered, and dynamic environments requires a planner that knows where and when danger will appear, not only that a trajectory is dangerous. One-stage learning-based planners trained with differentiable privileged costs are fast and expert-free, but the only signal reaching their encoder is a scalar trajectory cost with no spatial or temporal structure, so avoidance degrades into late reactive maneuvers. We present RiskFly, a one-stage planner that predicts risk in the same space in which it acts. A dual-stream observation pairs a short depth sequence with a frustum-aligned inverted spherical range-map sequence, whose angular cells match the end-state proposals one to one. An auxiliary head regresses a frustum-aligned spatio-temporal risk field, supervised by a privileged closest-point-of-approach (CPA) target. This self-predicted field is queried differentiably along the instantiated quintic trajectory at its own arrival times, and also enters the training objective, so representation supervision and planning gradients meet in a single space. Privileged signals are discarded at deployment, and the planner runs map-free from onboard depth and proprioception. Extensive simulation and zero-shot real-world flights on resource-constrained platforms show higher success rates at comparable end-to-end latency.
\end{abstract}

 \section{Introduction}
Autonomous navigation is an essential capability for unmanned aerial vehicles (UAVs)~\cite{oleynikova2020open}. Limited onboard computation and low-fidelity sensing make high-speed collision-free flight in cluttered and dynamic environments demanding~\cite{zhou2021ego}, since the vehicle must decide continuously under motion blur, a narrow field of view (FoV), and frequent occlusions~\cite{song2023learning}. Classical modular pipelines of perception, mapping, planning, and control remain interpretable~\cite{zhou2019robust,ren2022bubble}. However, their detection and prediction modules add latency and accumulate error precisely when the reaction window is shortest~\cite{xu2023real,lu2025fapp}. Learning-based planners instead unify perception and planning in a single network, and infer actions directly from raw observations~\cite{lu2024you,fan2025obsnet,xu2026p2m,xu2025navrl}.

Among these, the one-stage paradigm~\cite{lu2024you} is attractive for embedded deployment. The FoV is partitioned into a frustum grid, a network regresses one scored terminal-state proposal per cell, and a feasible polynomial trajectory follows in closed form. Because the trajectory is an analytic differentiable function of the network output, the policy can be trained without expert demonstrations, using a physically grounded cost from privileged simulation geometry, and the natural extension to dynamic scenes feeds a short depth sequence to a temporal encoder and adds a dynamic-avoidance cost derived from ground-truth actor states.

\begin{figure}[t]
    \vspace{0.1cm}
	\centering
	\includegraphics[width=0.85\linewidth]{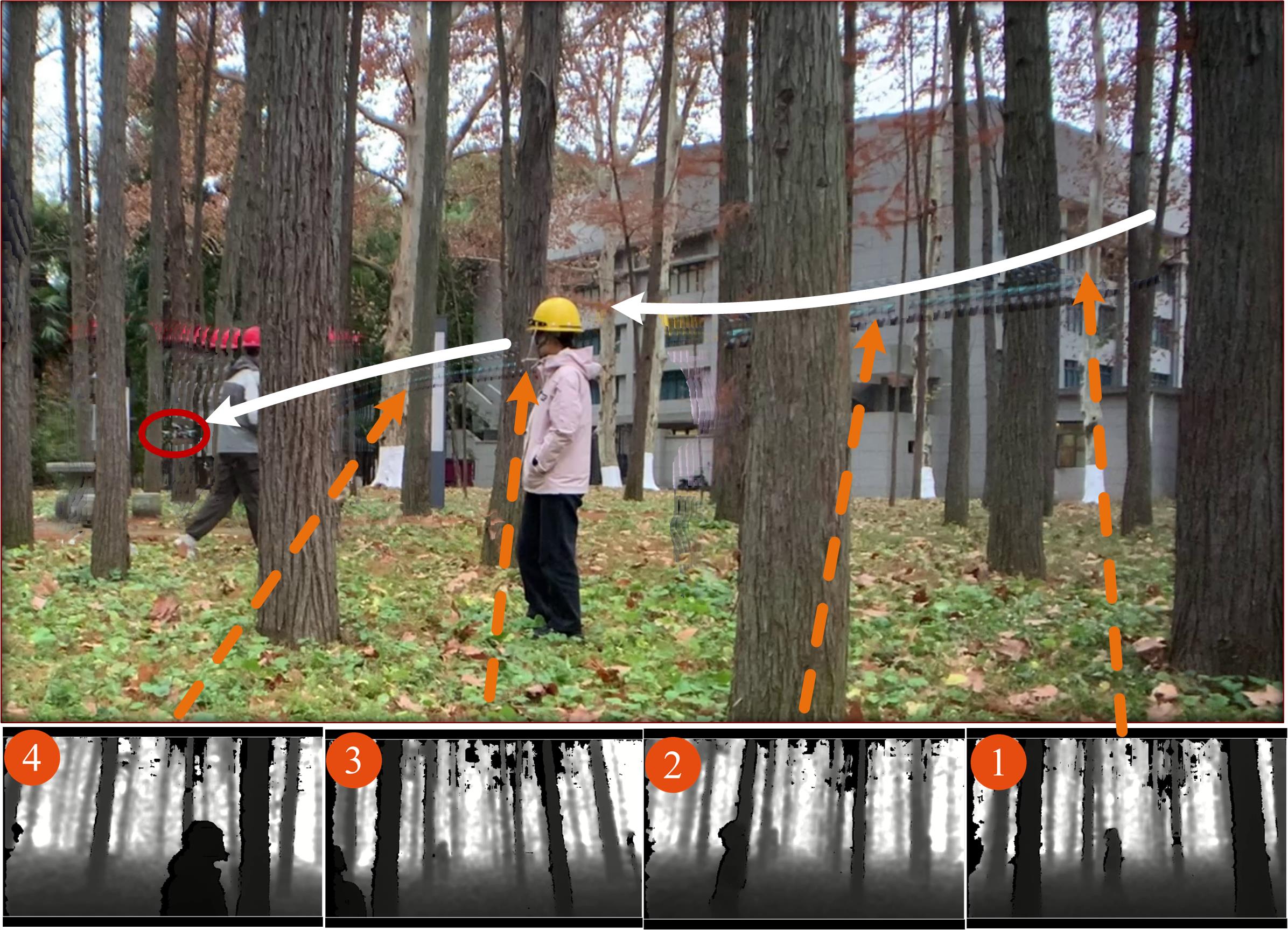}
	\caption{\textbf{Snapshot of real-world flight experiments.} Zero-shot sim-to-real deployment of the proposed one-stage planner in a previously unseen forest with two walking pedestrians, using only onboard depth and proprioception. White arrows give the flight direction and the red ellipse marks the quadrotor; the bottom strip shows the onboard depth frames $1$--$4$ taken at the instants the dashed lines indicate.}
	\label{fig:teaser}
\end{figure}

This design, however, leaves two structural gaps. First, the supervision is spatially blind. A dynamic cost evaluated on the instantiated trajectory is a scalar, so the encoder learns that some trajectory was risky, but not which FoV direction or which future instant caused the penalty; when the shared representation is not motion aware, these gradients are attributed to static structure. Second, the observation is not aligned with the decision space: raw depth measures range along the optical axis and is distorted toward the periphery, whereas the planner reasons on an angular frustum grid, so the network must absorb this mapping implicitly, which wastes capacity and widens the sim-to-real gap.

A third gap concerns when safety is evaluated. Existing risk representations are either instantaneous clearances~\cite{wang2026apf} or a time-collapsed azimuthal vector~\cite{mei2026arrfly}, yet a trajectory reaches a given bearing at a specific time $t_k$, and that is where its safety should be queried. Spatio-temporal planners exploit this observation~\cite{srivastava2026spot,chen2022rast}, but one-stage learned planners have not internalized it.

We therefore propose RiskFly, a one-stage planner that predicts risk in the same space in which it acts. Following~\cite{mei2026arrfly}, we build an inverted spherical range map from depth, align it to the frustum grid of the planner, and use it as one of two parallel streams alongside the raw depth sequence, combining an explicit geometry-consistent representation with dense implicit appearance cues. An auxiliary head regresses a frustum-aligned spatio-temporal risk field supervised by a privileged closest-point-of-approach (CPA) target, and this self-predicted field is queried along the instantiated trajectory at its own arrival times, so that representation supervision and planning gradients act on one shared field. All privileged quantities are training-only, and at inference the planner is map-free and detection-free.

Our contributions are summarized as follows:
\begin{enumerate}
    \item We propose a dual-stream observation for one-stage dynamic planning, pairing a short depth sequence with a frustum-aligned inverted spherical range-map sequence, so that the perceptual representation, the end-state proposals, and the risk prediction live on one angular grid.

    \item We propose a frustum-aligned spatio-temporal risk field supervised by a privileged CPA-based target. An arrival-time differentiable coupling then injects the self-predicted field into the trajectory objective and into proposal scoring.

    \item We validate the system on simulation benchmarks spanning static and dynamic density gradients and on zero-shot real-world deployment, with ablations isolating the three components.
\end{enumerate}

\section{Related Work}

\subsection{Object-Aware Dynamic Obstacle Avoidance}

Traditional UAV navigation in dynamic environments relies on explicit obstacle-motion models~\cite{xu2023real}. Hybrid representations model dynamic obstacles as geometric primitives while voxelizing the static scene, and an adaptive potential field can be driven from a spherical obstacle map~\cite{wang2026apf}. Spatio-temporal formulations instead model time-varying occupancy directly: risk-aware corridors grow from particle-based dynamic maps~\cite{chen2022rast}, while SPOT searches a space-time graph so that collision checking happens when the vehicle actually arrives~\cite{srivastava2026spot}. At the control level, model predictive control (MPC) handles perception uncertainty and kinodynamic constraints, from intent prediction~\cite{xu2025intent} and perception-enhanced control~\cite{liu2025dynamic} to online covariance adaptation~\cite{lu2025fapp,lim2026safe}.

These pipelines are interpretable, but depend on precise state estimation of every moving object: detection, association, and filtering form a serial dependency whose latency and error grow with occlusion and obstacle count, which forces conservative speeds and motivates methods that reason about motion from raw sensing.

\subsection{Learning-Based Dynamic Obstacle Avoidance}

Perception-to-action policies reduce decision latency by mapping observations directly to trajectory parameters. Early planners extract geometric cues from single-frame depth: YOPO~\cite{lu2024you} regresses frustum-conditioned end states with guidance learning, inverse reinforcement learning achieves agile flight in static clutter~\cite{kim2025rapid}, and mapless MPC with dual KD-trees keeps an optimizer in the loop without a map~\cite{zhang2025mapless}. Such snapshot-based policies generalize poorly to dynamic scenes, because ego-motion-induced depth variation masks true obstacle motion.

Recent work therefore injects temporal information: aggregating historical range scans into a compact angular map captures obstacle kinematics~\cite{fan2025obsnet}, a point-flow representation makes relative motion explicit~\cite{xu2026p2m}, and RL on point clouds achieves low-latency control~\cite{xu2025flying}. Others retain an explicit detector with a safety shield~\cite{xu2025navrl}, learn a continuous dynamic distance map~\cite{zhang2026e2e}, or address multi-modality with conditional diffusion~\cite{wang2026aerodiffusion}. Most are nevertheless shaped by aggregated scalar rewards, and internalize late reactive evasions rather than anticipatory detours. A risk-guided reinforcement-learning planner supervises a spatio-temporal encoder with a privileged CPA risk map~\cite{mei2026arrfly}, but the risk is compressed into an azimuthal vector and detached before the actor, so no gradient links risk to the executed motion.

\subsection{Field Representations and Differentiable Guidance}

A complementary line of work rewrites discrete planning decisions as continuous scalar fields whose gradients are consumed directly by an optimizer. Time-of-arrival geodesic fields replace Euclidean goal attraction so that guidance flows around occluders~\cite{liu2026toa}, cloth simulation fields delineate traversable space from a depth matrix~\cite{xing2025csf}, and optical-flow fields serve as the policy input and encode the approach rate~\cite{hu2025flow,deng2026fpv}. These work because a dense differentiable signal is a far richer training target than a scalar, but they are defined over space alone.

We take this step. RiskFly keeps the expert-free differentiable one-stage paradigm of~\cite{lu2024you}, adopts the explicit range-map abstraction of~\cite{mei2026arrfly} but aligns it to the planning frustum, and turns the privileged risk signal into a spatio-temporal field queried at the arrival times of the trajectory.

\section{Proposed Approach}
    \begin{figure*}[!thp]
        \centering
        \includegraphics[width=0.85\linewidth]{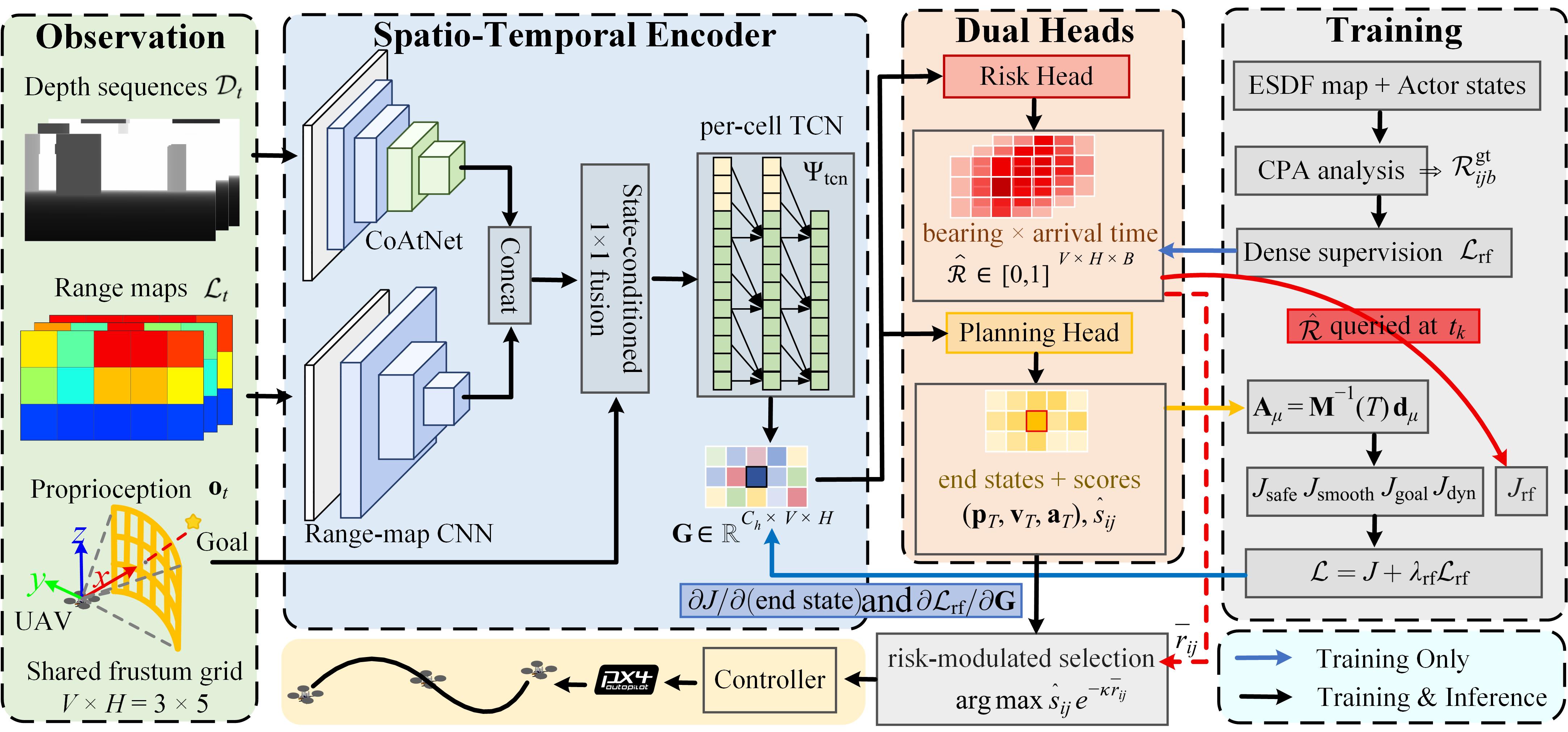}
        \caption{{\bf System overview.} A depth sequence $\mathcal{D}_t$, a frustum-aligned range-map sequence $\mathcal{L}_t$, the proprioceptive state $\mathbf{o}_t$ and the goal direction are encoded on the shared $V\times H$ frustum grid: a CoAtNet branch and a range-map CNN are concatenated, fused with the state by a $1\times1$ convolution, and aggregated by a per-cell TCN into $\mathbf{G}$. A \emph{risk head} regresses the spatio-temporal risk field $\hat{\mathcal{R}}\in[0,1]^{V\times H\times B}$ indexed by bearing and arrival time, and a \emph{planning head} regresses one scored terminal-state proposal per cell. During training, the ESDF and ground-truth actor states supply the CPA target for $\mathcal{L}_{\text{rf}}$ and the trajectory costs, while $\hat{\mathcal{R}}$ is queried along the instantiated trajectory \emph{at its own arrival times} as $J_{\text{rf}}$. At inference the privileged modules are detached and $\bar{r}_{ij}$ modulates proposal selection.}
        \label{fig:overview}
    \end{figure*}

\subsection{System Overview}
As shown in Fig.~\ref{fig:overview}, at every planning cycle the network maps a dual-stream observation and the proprioceptive state to a frustum-aligned spatio-temporal risk field and a set of scored body-frame terminal states. During training, each terminal state is instantiated as a feasible quintic trajectory, and a differentiable objective enforces static and dynamic safety, goal reaching, smoothness, and consistency with the self-predicted field. At inference, only onboard depth and proprioception are needed, and the planner emits a trajectory in closed form without mapping, detection, or tracking.

\subsection{End-State Prediction with a One-Stage Planner}
\label{sec:endstate}
Following~\cite{lu2024you}, we formulate local navigation as end-state prediction over a fixed receding horizon. All states, observations, and outputs are expressed in the UAV body frame.

\subsubsection{Trajectory Representation}
The local trajectory is represented as three independent quintic polynomials, one per axis $\mu\in\{x,y,z\}$:
\begin{equation}
p_{\mu}(t) = \sum_{k=0}^{5} a_{\mu,k} t^{k}, \quad t \in [0, T].
\label{eq:poly}
\end{equation}
With a fixed horizon $T$, the coefficient vector $\mathbf{A}_{\mu} = [a_{\mu,0}, \ldots, a_{\mu,5}]^{\top}$ is uniquely determined by the boundary conditions $\mathbf{d}_{\mu}=[p_{\mu}(0),\dot{p}_{\mu}(0),\ddot{p}_{\mu}(0),p_{\mu}(T),\dot{p}_{\mu}(T),\ddot{p}_{\mu}(T)]^{\top}$, which yields the closed-form mapping
\begin{equation}
\mathbf{A}_{\mu} = \mathbf{M}^{-1}(T)\,\mathbf{d}_{\mu},
\label{eq:coef_map}
\end{equation}
where $\mathbf{M}(T)$ is shared across axes and $\mathbf{M}^{-1}(T)$ is precomputed offline. Since $p(0)=0$ in the body frame, planning reduces to selecting the terminal state, and this differentiable map lets a physically grounded cost be back-propagated to the network output without expert labels.

\subsubsection{Frustum-Conditioned End-State Proposals}
To address the multi-modal nature of obstacle avoidance, the FoV is divided into a frustum grid of $V \times H$ cells, with $V=3$ and $H=5$, and the network predicts one scored terminal-state proposal per cell $(i,j)$. The body-frame velocity $\mathbf{v}_b$, acceleration $\mathbf{a}_b$, and goal direction $\mathbf{g}_b$ are normalized by $(v_{\max},a_{\max})$ and unit length, rotated by the precomputed body-to-cell rotation $\mathbf{R}_{ij}$, and stacked into a 9D cell-conditioned state $\mathbf{o}_t$.

A shared head then outputs a 10D vector per cell: $(u_{\psi},u_{\theta},u_{d})$ parameterize the terminal position in yaw, pitch, and range, $\mathbf{u}_{v},\mathbf{u}_{a}\in\mathbb{R}^{3}$ the terminal velocity and acceleration, and $u_{s}$ is a score. Scaled $\tanh$ squashing bounds $\hat{\psi},\hat{\theta}$ to the angular span of the cell and $\hat{d}\in[0,d_{\max}]$, while $\mathbf{u}_{v},\mathbf{u}_{a}$ are bounded element-wise by $(v_{\max},a_{\max})$, so kinematic limits hold by construction.

\subsubsection{Risk-Modulated Proposal Selection}
Unlike~\cite{lu2024you}, the score is not the only criterion. With $\bar{r}_{ij}=\max_{b}\hat{\mathcal{R}}_{ijb}$ the peak predicted risk of cell $(i,j)$ over the horizon (Sec.~\ref{sec:risk}), we select
\begin{equation}
(i^{*},j^{*}) = \arg\max_{i,j}\ \Big[\,\hat{s}_{ij}\,\exp\!\big(-\kappa\,\bar{r}_{ij}\big)\Big],
\label{eq:score_mod}
\end{equation}
where $\hat{s}_{ij}=\mathrm{softplus}(u_{s,ij})$ and $\kappa>0$ controls how strongly anticipated risk suppresses a direction. Because $\bar{r}_{ij}$ and $\hat{s}_{ij}$ live on the same grid, a cell predicted to become hazardous within the horizon is demoted even if its current geometric clearance is large. The selected cell is decoded to a terminal state and mapped to quintic coefficients via~\eqref{eq:coef_map}.

\subsection{Dual-Stream Frustum-Aligned Perception}
\label{sec:network}
Single-frame policies cannot disentangle obstacle motion from ego-motion-induced depth change, and a single raw-depth stream must learn the mapping from optical-axis pixels to the angular decision space implicitly. We therefore adopt a dual-stream frustum-aligned observation.

\begin{figure}[t]
    \vspace{0.1cm}
    \centering
    \includegraphics[width=0.95\linewidth]{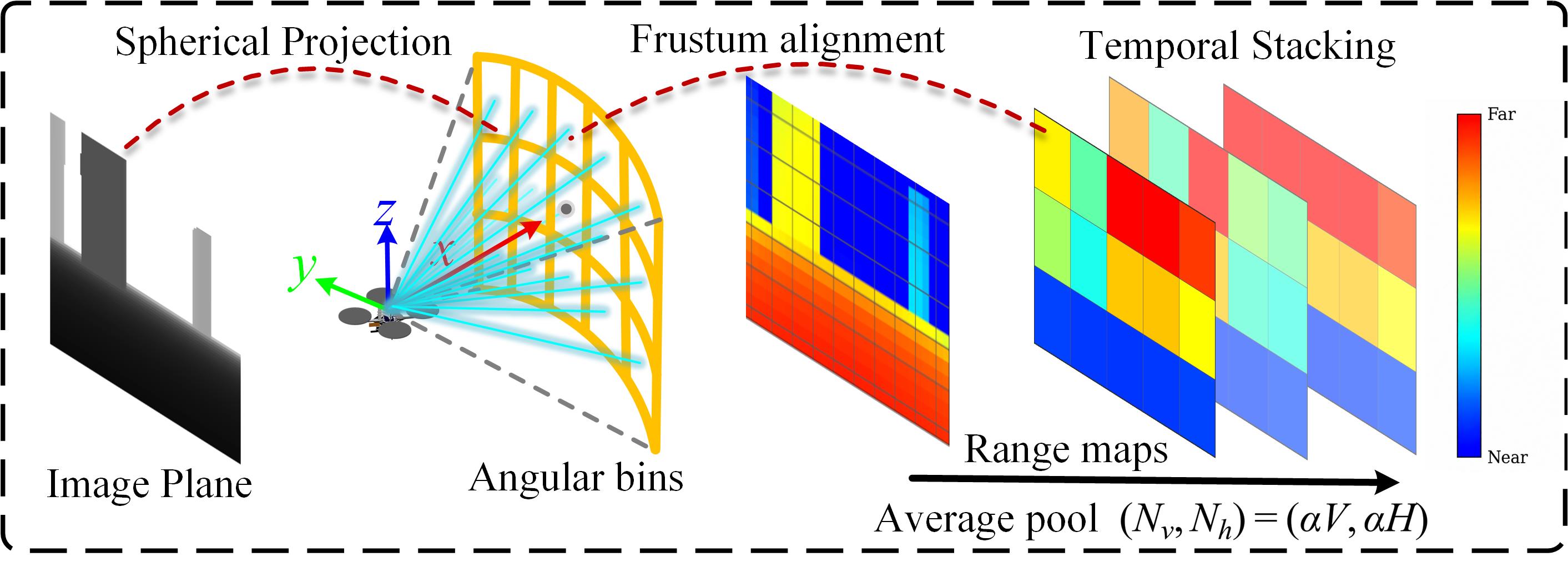}
    \caption{\textbf{Construction of the frustum-aligned range map.} Min-pooled depth is spherically projected onto $N_h\times N_v$ angular bins that subdivide the planning frustum, and optical-axis depths are converted to Euclidean ranges and inverted so that proximity yields a high response. Average pooling with $(N_v,N_h)=(\alpha V,\alpha H)$ maps the result exactly onto the $V\!\times\!H$ grid, and the $L_r$ most recent maps are stacked into $\mathcal{L}_t$.}
    \label{fig:rangemap}
\end{figure}

\subsubsection{Explicit frustum range maps}
Following~\cite{mei2026arrfly}, the depth image is min-pooled with a $3\times3$ window into a conservatively filtered map $D^{\min}_{t}$, with invalid pixels recovered by interpolation. As shown in Fig.~\ref{fig:rangemap}, the valid FoV is discretized into $N_h$ azimuth bins $\theta_i$ and $N_v$ elevation bins $\phi_j$, and optical-axis depths are converted into Euclidean ranges and inverted, so that proximity yields a high response:
\begin{equation}
L_{t,ij} = \max\!\left(0,\; d_{\max} - \tilde{D}_{t,ij}\sqrt{1+\tan^{2}\theta_i+\tan^{2}\phi_j}\right),
\label{eq:rangemap}
\end{equation}
where $\tilde{D}_{t,ij}$ is sampled from $D^{\min}_{t}$ and $d_{\max}$ is the maximum reliable sensing range. Unlike~\cite{mei2026arrfly}, we choose $(N_v,N_h)$ as an integer multiple of the planning grid, $(N_v,N_h)=(\alpha V,\alpha H)$ with $\alpha=6$, so that a single average pooling maps $L_t$ exactly onto the $V\times H$ frustum partition and perception, risk prediction, and action proposals share one angular grid by construction. A stack of the $L_r$ most recent range maps, $\mathcal{L}_t=\{L_{t-L_r+1},\ldots,L_t\}$, forms the explicit stream.

\subsubsection{Implicit depth stream}
The range map is coarse and conservative, which makes it domain-robust but discards fine structure such as thin branches and occluded silhouettes. We therefore retain the raw depth sequence $\mathcal{D}_t=\{\mathbf{D}_{t-L+1},\ldots,\mathbf{D}_t\}$ with $L=6$, encoded frame-wise by a lightweight CoAtNet backbone~\cite{dai2021coatnet} whose final stage is aligned to the frustum partition, $\mathbf{F}_{t-\ell}=\Phi_{\text{CoAt}}(\mathbf{D}_{t-\ell})\in\mathbb{R}^{C\times V\times H}$.

\subsubsection{State-conditioned temporal fusion}
Per-frame features of the two streams are concatenated per cell, $\mathbf{Z}_{t-\ell}=[\mathbf{F}_{t-\ell},\,\Phi_{\text{rm}}(L_{t-\ell})]$. The proprioceptive vector $\mathbf{o}_t$ is embedded by a small MLP, broadcast to all cells, and fused with $\mathbf{Z}_{t-\ell}$ through a $1{\times}1$ convolution, so the network can attribute inter-frame range change to ego-motion rather than to obstacle motion. A per-cell temporal convolutional network (TCN) then aggregates the sequence into a motion-aware map $\mathbf{G}=\Psi_{\text{tcn}}(\tilde{\mathbf{Z}}_{t-L+1},\ldots,\tilde{\mathbf{Z}}_{t})\in\mathbb{R}^{C_h\times V\times H}$. Both heads read from $\mathbf{G}$, and the planning head additionally receives a late state shortcut by concatenating $\mathbf{o}_t$ at each cell.

\subsection{Privileged Spatio-Temporal Risk Field}
\label{sec:risk}
\begin{figure}[!t]
    \centering
    \includegraphics[width=0.85\linewidth]{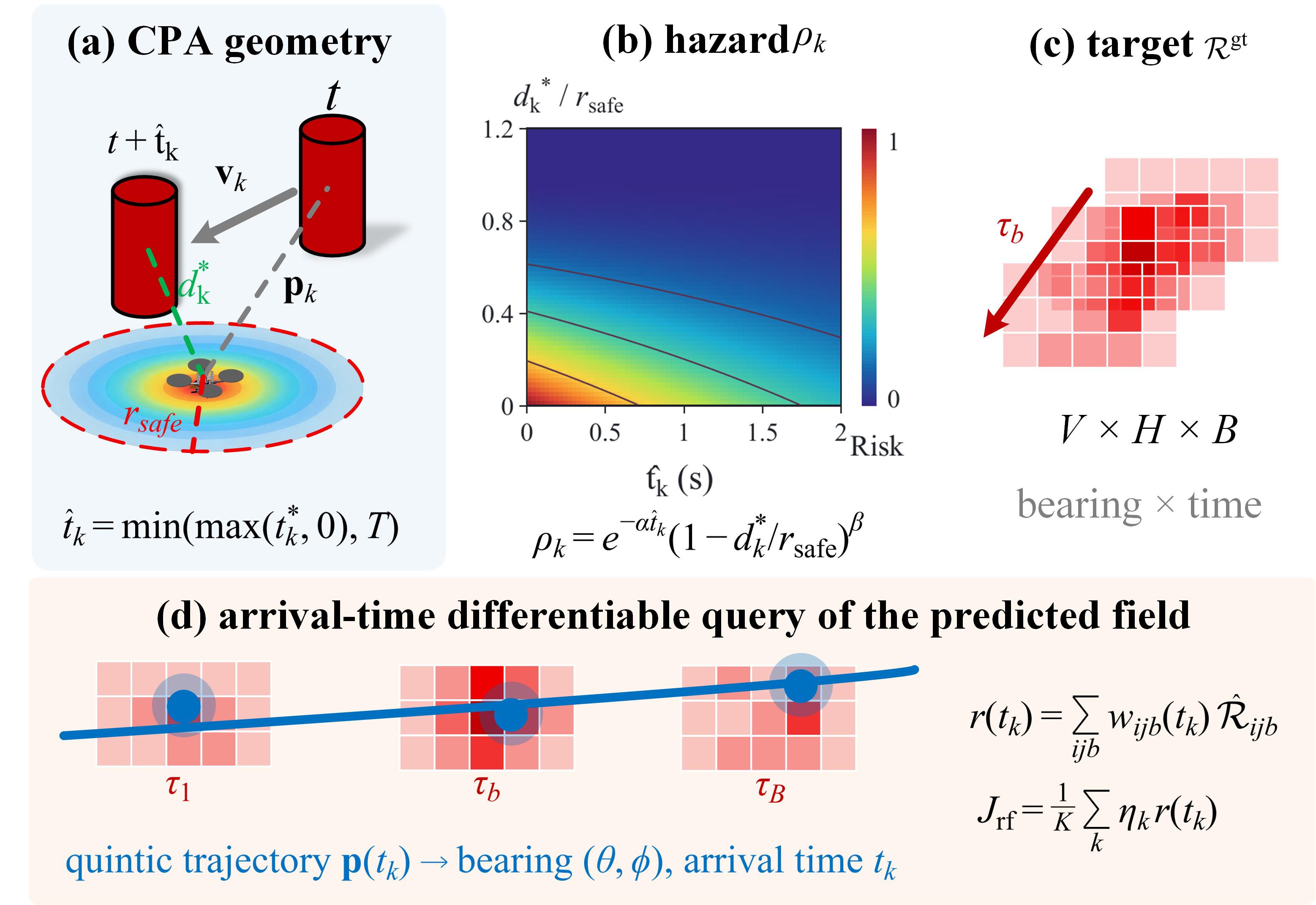}
    \caption{{\bf Spatio-temporal risk field and arrival-time query.} (a) The CPA of the relative motion defines $\hat{t}_k$ and $d^{*}_k$. (b) A bivariate hazard maps $(\hat{t}_k,d^{*}_k/r_{\text{safe}})$ to a continuous risk score. (c) Hazards are rasterized onto the frustum-time grid $\mathcal{R}^{\text{gt}}\in[0,1]^{V\times H\times B}$, indexed by bearing and arrival time. (d) The instantiated trajectory queries the \emph{predicted} field at its own bearing and arrival time, so the cost is evaluated where and when the vehicle will actually be.}
    \label{fig:riskfield}
\end{figure}

The core of RiskFly is a dense training target with structure in both space and time, telling the encoder where and when a collision is being set up.

\subsubsection{CPA-based risk ground truth}
For the $k$-th actor, let $\mathbf{p}_k,\mathbf{v}_k$ be its ego-centric position and velocity relative to the UAV, taken from privileged simulator states. Under an approximately constant relative velocity over the horizon $T$, the time of closest approach is
\begin{equation}
t^{*}_k=-\frac{\mathbf{p}_k^{\top}\mathbf{v}_k}{\|\mathbf{v}_k\|^{2}+\epsilon},\qquad
\hat{t}_k=\min\!\big(\max(t^{*}_k,0),\,T\big),
\label{eq:cpa}
\end{equation}
with $\epsilon=10^{-5}$ for numerical stability. The corresponding minimum separation is $d^{*}_k=\|\mathbf{p}_k+\hat{t}_k\mathbf{v}_k\|$. The hazard of actor $k$ is a bivariate function of temporal urgency and spatial proximity,
\begin{equation}
\rho_k=
\begin{cases}
e^{-\alpha \hat{t}_k}\Big(1-\dfrac{d^{*}_k}{r_{\text{safe}}}\Big)^{\beta}, & \hat{t}_k\ge 0,\ d^{*}_k<r_{\text{safe}}\\[4pt]
0, & \text{otherwise,}
\end{cases}
\label{eq:hazard}
\end{equation}
where $r_{\text{safe}}=r_{\text{drone}}+r_{\text{obs}}+d_{\text{buffer}}$. Setting $\beta=2$ ensures $C^1$ smoothness at the safety boundary.

\subsubsection{From a vector to a field}
Prior work pools these hazards over elevation into a one-dimensional azimuthal vector~\cite{mei2026arrfly}, discarding the vertical escape direction and the time at which the hazard occurs. We instead rasterize them onto a frustum-time grid: the horizon is split into $B$ uniform bins with centers $\tau_b$, and each actor is assigned to the cells its safety sphere $r_{\text{safe}}$ intersects. The target field is
\begin{equation}
\mathcal{R}^{\text{gt}}_{ijb}=\max_{k\in\mathcal{A}_{ij}}\ \rho_k\,\exp\!\left(-\frac{(\hat{t}_k-\tau_b)^{2}}{2\sigma_\tau^{2}}\right),
\label{eq:risk_gt}
\end{equation}
where $\mathcal{A}_{ij}$ is the set of actors intersecting cell $(i,j)$ and $\sigma_\tau$ spreads each hazard over neighbouring bins, keeping the target smooth under small prediction-time errors. The field $\mathcal{R}^{\text{gt}}\in[0,1]^{V\times H\times B}$ thus states, for every bearing and future instant, how dangerous that direction will be, which is exactly what a trajectory implicitly asks when it schedules a waypoint.

\subsubsection{Risk head and supervision}
An auxiliary convolutional head decodes $\mathbf{G}$ into $\hat{\mathcal{R}}=f_{\text{risk}}(\mathbf{G})\in[0,1]^{V\times H\times B}$, supervised by
\begin{equation}
\mathcal{L}_{\text{rf}}=\frac{1}{VHB}\sum_{i,j,b}w_{ijb}\big(\hat{\mathcal{R}}_{ijb}-\mathcal{R}^{\text{gt}}_{ijb}\big)^{2},
\label{eq:risk_loss}
\end{equation}
with $w_{ijb}=1+\gamma\,\mathcal{R}^{\text{gt}}_{ijb}$ up-weighting the sparse hazardous entries. Since $\hat{\mathcal{R}}$ is decoded from the feature map that also feeds the planning head, this dense signal shapes the shared representation directly, not only through a scalar cost.

\subsection{Risk-Coupled Differentiable Training Objective}
\label{sec:cost}
Realistic dynamic-scene costs are non-differentiable and label regression is misaligned with collision risk, so we instantiate each terminal state into a quintic trajectory and minimize a differentiable total cost:
\begin{equation}
J = \lambda_M J_{\text{smooth}} + \lambda_S J_{\text{safe}} + \lambda_G J_{\text{goal}} + \lambda_D J_{\text{dyn}} + \lambda_R J_{\text{rf}},
\label{eq:total_cost}
\end{equation}
trained jointly with the auxiliary term as $\mathcal{L}=J+\lambda_{\text{rf}}\mathcal{L}_{\text{rf}}$.

\subsubsection{Static safety, smoothness, and goal guidance}
The horizon is discretized into $K$ timestamps $\{t_k\}_{k=1}^{K}$; querying the privileged ESDF at $\mathbf{p}(t_k)$ gives signed distances $d_k$, and an exponential barrier penalizes insufficient clearance:
\begin{equation}
J_{\text{safe}} = \frac{1}{K}\sum_{k=1}^{K}\exp\!\left(-\frac{d_k-d_0}{r}\right),
\label{eq:safety_cost}
\end{equation}
with margin $d_0$ and decay scale $r$. Smoothness penalizes the time integrals of squared jerk and acceleration, which are closed-form quadratics in the boundary derivatives, $J_{\text{smooth}} = \sum_{\mu}\mathbf{d}_{\mu}^{\top}(\mathbf{R}_J+\omega_a\mathbf{R}_A)\mathbf{d}_{\mu}$ with constant Hessians. Goal guidance combines direction, terminal-velocity alignment, and a progressively weighted position term, $J_{\text{goal}} = J_{\text{dir}} + \omega_v J_{\text{vel}} + \omega_p J_{\text{pos}}$, as in the one-stage formulation.

\subsubsection{Dynamic safety from privileged actor states}
As a geometric complement to the learned field, we retain a direct dynamic barrier. Each actor is a vertical cylinder inflated by a time-decaying margin $m(t_k)=m_{\max}(1-t_k/T)$ and rolled out under a constant-acceleration model scaled by speed hypotheses $h\in\mathcal{H}$. With $\delta_{i,h}(t_k)$ the signed clearance to the inflated surface,
\begin{equation}
J_{\text{dyn}} = \sum_{h\in\mathcal{H}} w_h\, \frac{1}{K}
\sum_{k=1}^{K}
\exp\!\left(-\frac{\min_i \delta_{i,h}(t_k)-d_0}{r}\right),
\label{eq:dyn_cost}
\end{equation}
sharing $d_0$ and $r$ with~\eqref{eq:safety_cost}.

\subsubsection{Arrival-time risk coupling}
The remaining term closes the loop. Rather than detaching $\hat{\mathcal{R}}$, we query the field of the network itself along the trajectory, at the time the vehicle reaches each sample. For $\mathbf{p}(t_k)$ we compute the bearing $(\theta(t_k),\phi(t_k))$ and a differentiable soft assignment over the frustum-time grid,
\begin{equation}
w_{ijb}(t_k)\!=\!\frac{\exp\!\big(\!-\tfrac{\Delta\theta_i^{2}}{2\sigma_\theta^{2}}\!-\!\tfrac{\Delta\phi_j^{2}}{2\sigma_\phi^{2}}\!-\!\tfrac{(t_k-\tau_b)^{2}}{2\sigma_\tau^{2}}\big)}{\sum_{i'\!,j'\!,b'}\exp\!\big(\!-\tfrac{\Delta\theta_{i'}^{2}}{2\sigma_\theta^{2}}\!-\!\tfrac{\Delta\phi_{j'}^{2}}{2\sigma_\phi^{2}}\!-\!\tfrac{(t_k-\tau_{b'})^{2}}{2\sigma_\tau^{2}}\big)},
\label{eq:soft_assign}
\end{equation}
with $\Delta\theta_i=\theta(t_k)-\theta_i$ and $\Delta\phi_j=\phi(t_k)-\phi_j$. The queried risk and the resulting cost are
\begin{equation}
r(t_k)=\!\!\sum_{i,j,b}\! w_{ijb}(t_k)\,\hat{\mathcal{R}}_{ijb},
\qquad
J_{\text{rf}}=\frac{1}{K}\sum_{k=1}^{K}\eta_k\, r(t_k),
\label{eq:risk_cost}
\end{equation}
where $\eta_k=\|\mathbf{p}(t_k)\|/d_{\max}$ down-weights samples very close to the vehicle, whose safety is already governed by~\eqref{eq:safety_cost}.

This coupling is more than a re-weighting of $J_{\text{dyn}}$. Gradients flow through $w_{ijb}(t_k)$ into the terminal state, pushing the planner toward bearings and arrival schedules that perception itself deems safe, and through $\hat{\mathcal{R}}$, so under-predicting risk along a chosen trajectory is penalized twice. Indexing by time further separates a bearing that is dangerous now from one that becomes dangerous only after the vehicle has passed.

At deployment, $\mathcal{R}^{\text{gt}}$, the ESDF, and all actor states are discarded, while $\hat{\mathcal{R}}$ is still produced by the retained risk head and consumed by~\eqref{eq:score_mod}, so the anticipatory behaviour remains available onboard at no extra sensing cost. Training is end-to-end with AdamW, and replanning at $50$~Hz compensates online for residual errors of~\eqref{eq:cpa}.

\section{Experiment}
 \begin{figure*}[!t]
    \centering
    \includegraphics[width=0.9\linewidth]{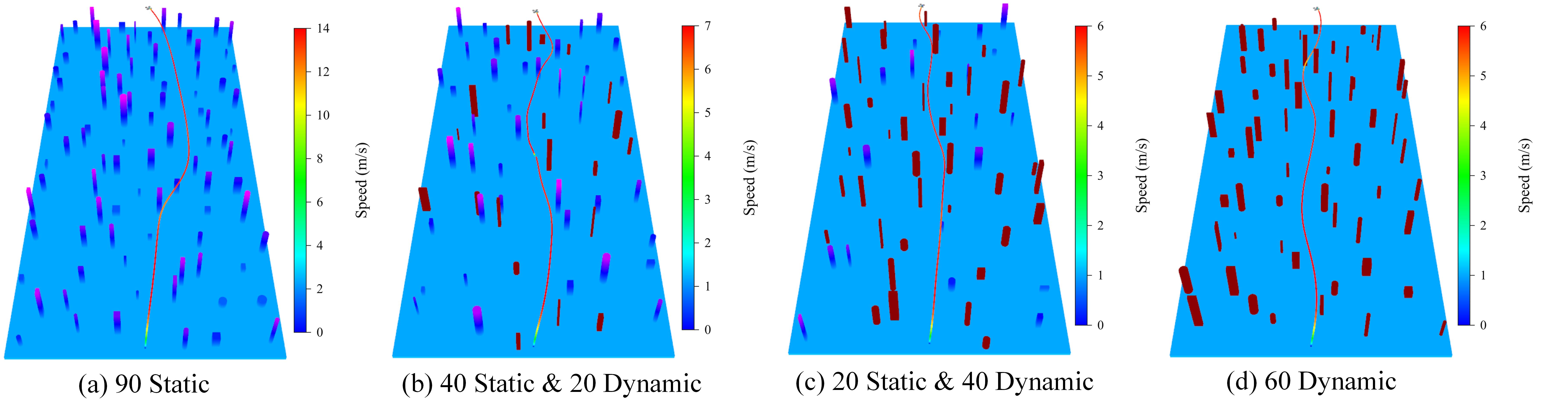}
    \caption{{\bf Simulation environments and performance of our method.} Four $30\times 60\,\mathrm{m}$ scenarios keep the obstacle count comparable while shifting the static-to-dynamic ratio: (a) 90 Static, (b) 40 Static \& 20 Dynamic, (c) 20 Static \& 40 Dynamic, (d) 60 Dynamic. Trajectories are color-coded by UAV speed ($\mathrm{m/s}$), with a scale that differs across scenarios. Static obstacles are magenta; dynamic obstacles are red and move randomly at $[1,4]\,\mathrm{m/s}$.}
    \label{fig:gazebo_sim}
\end{figure*}

\begin{table}[t]
    \setlength{\tabcolsep}{3pt}
    \footnotesize
    \centering
    \caption{Baseline Comparison in Cluttered Environments with Different Densities of Static and Dynamic Obstacles}
    \begin{threeparttable}
    {
    \begin{tabular}{m{0.95cm}<{\centering}m{1.95cm}<{\centering}m{0.6cm}<{\centering}m{0.95cm}<{\centering}m{0.95cm}<{\centering}m{0.95cm}<{\centering}m{0.72cm}<{\centering}}
       \toprule[1.2pt]
       \textbf{Scene} & \textbf{Method} & \textbf{$\eta(\%)$} & \textbf{$v_a(m/s)$} & \textbf{$v_m(m/s)$} & \textbf{$t_p(ms)$} & \textbf{$R_l$}\\
       \midrule[0.8pt]

    \multirow{5.5}{*}{\rotatebox[origin=c]{90}{90 Static}}
     & Intent-MPC & 80 & 1.79 & 2.21 & 48.32 & 1.15\\
     & NavRL & 85 & 1.83 & 2.41 & 34.68 & 1.17\\
     & Avoid-MPC & 76 & \secondbest{6.51} & \secondbest{8.42} & 21.74 & \best{1.04}\\
     & YOPO & \secondbest{89} & 5.41 & 6.38 & \best{4.29} & \secondbest{1.06}\\
     & Proposed & \best{94} & \best{10.46} & \best{13.95} & \secondbest{7.38} & 1.07\\
     \midrule

     \multirow{5.5}{*}{\rotatebox[origin=c]{90}{\shortstack{40 Static \&\\20 Dynamic}}}
     & Intent-MPC & 52 & 1.61 & 2.17 & 52.41 & 1.23\\
     & NavRL & \secondbest{76} & 1.68 & 2.46 & 37.92 & 1.26\\
     & Avoid-MPC & 50 & \best{6.87} & \best{8.63} & 22.05 & \best{1.03}\\
     & YOPO & 74 & 4.85 & 6.09 & \best{4.34} & \secondbest{1.05}\\
     & Proposed & \best{89} & \secondbest{5.46} & \secondbest{6.92} & \secondbest{7.41} & 1.06\\
     \midrule

     \multirow{5.5}{*}{\rotatebox[origin=c]{90}{\shortstack{20 Static \&\\40 Dynamic}}}
     & Intent-MPC & 36 & 1.52 & 2.14 & 55.18 & 1.37\\
     & NavRL & \secondbest{58} & 1.55 & 2.48 & 40.46 & 1.43\\
     & Avoid-MPC & 34 & \best{6.75} & \best{8.36} & 22.48 & \best{1.08}\\
     & YOPO & 48 & 4.66 & \secondbest{6.13} & \best{4.41} & \secondbest{1.11}\\
     & Proposed & \best{78} & \secondbest{4.73} & 5.89 & \secondbest{7.62} & 1.12\\
     \midrule

     \multirow{5.5}{*}{\rotatebox[origin=c]{90}{60 Dynamic}}
     & Intent-MPC & 26 & 1.39 & 2.11 & 57.64 & 1.41\\
     & NavRL & \secondbest{44} & 1.49 & 2.43 & 42.36 & 1.49\\
     & Avoid-MPC & 20 & \best{6.98} & \best{8.61} & 22.63 & \best{1.06}\\
     & YOPO & 38 & \secondbest{4.74} & 5.87 & \best{4.52} & \secondbest{1.13}\\
     & Proposed & \best{67} & 4.58 & \secondbest{5.96} & \secondbest{7.69} & 1.14\\
     \bottomrule[1.2pt]
    \end{tabular}
    }
    \end{threeparttable}
    \label{tbl:baseline_compare}
\end{table}

\subsection{Implementation Details}
Simulation is performed in Gazebo on Ubuntu 20.04 with an Intel i7-13700KF CPU and an NVIDIA RTX 3090 Ti GPU. For real-world experiments the same policy is deployed on two cost-effective onboard computers, a Rockchip RK3588 and an NVIDIA Jetson Orin NX, with state estimation from VINS-Fusion~\cite{qin2018vins}. The frustum grid is $V\times H=3\times5$, the range map is built at $(N_v,N_h)=(18,30)$, the depth sequence length is $L=6$, and the risk field uses $B=5$ time bins over $T$. Privileged signals are used during training only.

\subsection{Simulation Experiment}
\subsubsection{Simulation Setup}\label{sec:sim_setup}
We benchmark against four representative baselines: the one-stage learning-based YOPO~\cite{lu2024you}, the intent-prediction-driven Intent-MPC~\cite{xu2025intent}, the RL-based NavRL~\cite{xu2025navrl}, and the mapless Avoid-MPC~\cite{zhang2025mapless}. The perception range is unified to $20\,\mathrm{m}$, and otherwise we follow the recommended configurations of the original papers. Our reference speed is $12\,\mathrm{m/s}$ in static scenes and $6\,\mathrm{m/s}$ in dynamic scenes as a soft target; YOPO uses $6$\,m/s and Avoid-MPC $8$\,m/s, while NavRL and Intent-MPC are designed for low-speed flight and use $v_{\max}=2.5\,\mathrm{m/s}$, so the comparison reflects the stable operating regime of each method rather than equal speeds.

The four $30\times 60\,\mathrm{m}$ scenarios of Fig.~\ref{fig:gazebo_sim} keep the total obstacle count comparable while shifting the static-to-dynamic ratio: (a) 90 Static, (b) 40 Static \& 20 Dynamic, (c) 20 Static \& 40 Dynamic, (d) 60 Dynamic. Dynamic obstacles move randomly at $[1,4]\,\mathrm{m/s}$. The success rate $\eta$ is computed from $50$ trials per scenario, and the average speed $v_a$, maximum speed $v_m$, perception-to-action latency $t_p$, and path efficiency $R_l = l/\|\boldsymbol{p}_g - \boldsymbol{p}_s\|$ from successful flights.

\subsubsection{Simulation Results and Analysis}\label{sec:result}
As summarized in Table~\ref{tbl:baseline_compare}, YOPO~\cite{lu2024you} achieves the lowest latency, $4.29$--$4.52\,\mathrm{ms}$, but its success rate falls from $89\%$ to $38\%$ as the dynamic fraction increases, consistent with its reliance on single-frame depth: moving obstacles are perceived as quasi-static, so avoidance is initiated late. Avoid-MPC~\cite{zhang2025mapless} attains the best path efficiency everywhere, $1.03$--$1.08$, yet collapses to $20\%$ in 60 Dynamic because its snapshot-based KD-tree perception assumes static obstacles. Intent-MPC~\cite{xu2025intent} and NavRL~\cite{xu2025navrl} survive the mildly dynamic scene, but their detection modules cost $34.68$--$57.64\,\mathrm{ms}$ and are sensitive to perceptual error, so Intent-MPC acts on stale forecasts and NavRL cannot replan in time, and both pay with detours that raise $R_l$ to $1.41$ and $1.49$.

Our method attains the highest success rate in all four scenes, and the margin over the strongest baseline widens from $5$ points in 90 Static to $23$ points in 60 Dynamic. Two effects are worth separating. The explicit range-map stream gives a conservative geometry-consistent encoding of proximal structure, which stabilizes the shared representation and lets the risk head converge on noisy depth; this is what allows the $10.46\,\mathrm{m/s}$ average and $13.95\,\mathrm{m/s}$ peak speed in the static scene, where the trajectory in Fig.~\ref{fig:gazebo_sim}(a) hugs the straight-line corridor and decelerates only at the gaps. The risk field supplies the anticipatory component, demoting a bearing that is open now but predicted to be occupied on arrival: as the dynamic fraction grows in Fig.~\ref{fig:gazebo_sim}(b)--(d), the speed profile becomes visibly more modulated and the vehicle slows before crossing traffic rather than after. The extra stream costs about $3\,\mathrm{ms}$ and raises $R_l$ to $1.14$, but the robustness and agility trade-off improves with scene complexity.

\begin{figure*}[!t]
    \centering
    \includegraphics[width=0.9\linewidth]{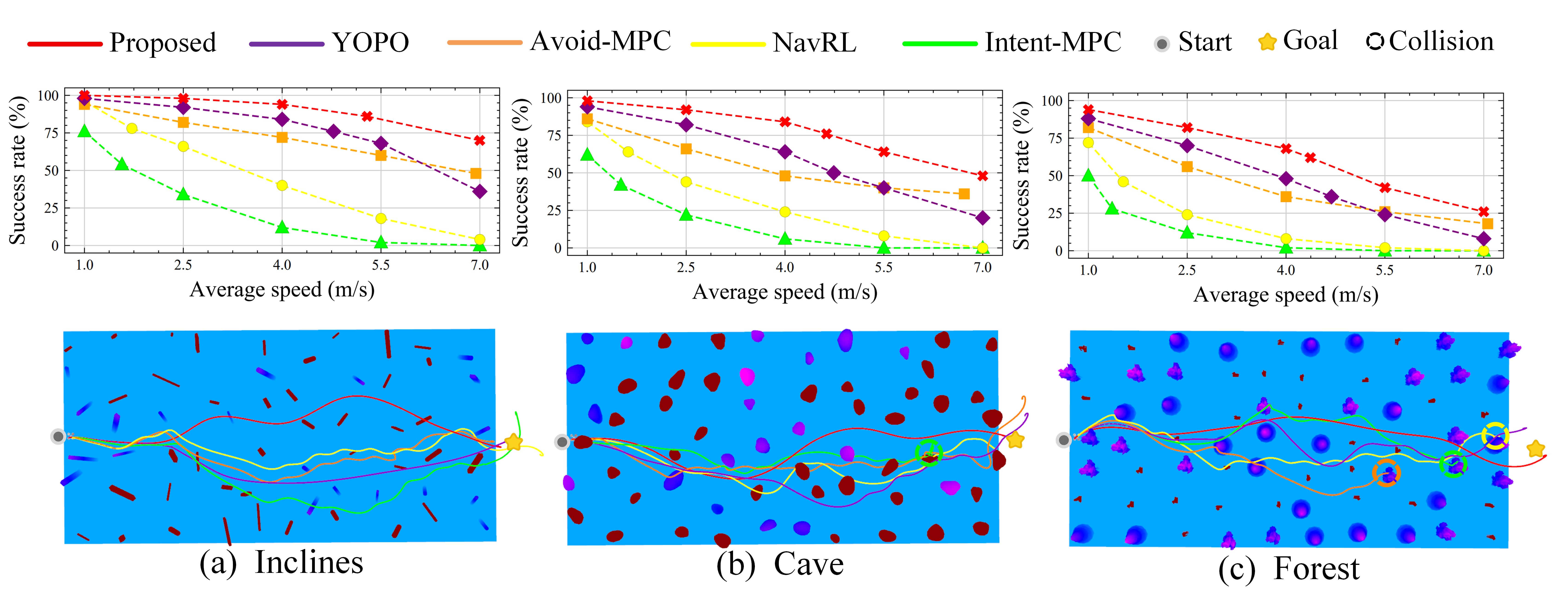}
    \caption{{\bf Generalization to unseen environments.} Three held-out environments, (a) Inclines, (b) Cave, (c) Forest, differ from the training scenes in geometry, obstacle shape, and free-space topology. Top: success rate over $50$ trials against the achieved average speed, obtained by sweeping the reference speed from $1.0$ to $7.0\,\mathrm{m/s}$. Bottom: trajectories of all five methods for one start--goal pair; grey dot is the start, yellow star the goal, dashed circles mark collisions. Dynamic obstacles are red, static obstacles blue to magenta.}
    \label{fig:unseen}
\end{figure*}

\subsubsection{Generalization to Unseen Environments}\label{sec:unseen}
To probe whether the learned risk field transfers rather than memorizes, we evaluate all five methods in three environments never seen during training (Fig.~\ref{fig:unseen}): Inclines of sloped slabs and thin tilted panels, Cave of densely packed blobs with narrow corridors, and Forest mixing tree-like canopies with low debris. Each mixes static and moving obstacles, and we sweep the reference speed and report the success rate over $50$ trials against the achieved average speed, so the comparison is made at equal realized agility.

The curves in the top row share the same ordering everywhere and separate with speed. At $4\,\mathrm{m/s}$ our method retains $94\%$, $84\%$, and $68\%$ in Inclines, Cave, and Forest, whereas the best baseline drops to $84\%$, $64\%$, and $48\%$; the margin is largest in Inclines, where planar surfaces make a bearing either safe for a long time or unsafe well before contact, exactly the structure the time-binned field represents. Intent-MPC and NavRL decay fastest and are unusable above $5.5\,\mathrm{m/s}$. The trajectories agree: our path leaves the direct corridor early and clears the slabs in one smooth arc, whereas the baselines commit to the low corridor and correct late, with one collision in Cave and three in Forest, and our paths are only $4$--$9\%$ longer than the shortest baseline path.

\subsection{Ablation Study}
We ablate three components under the settings of Sec.~\ref{sec:sim_setup}: w/o $J_{\text{rf}}$ keeps the risk head and its supervision but removes the arrival-time coupling, w/o ST-RF removes the risk head and its privileged supervision entirely, and w/o FRM removes the explicit range-map stream and doubles the depth-stream width. Results are summarized in Table~\ref{tbl:ablation_study}.

Removing the arrival-time coupling leaves the risk field accurate but inert: the encoder still learns where and when hazards form, yet nothing pushes the terminal state away from those bearings. Success drops by $10$, $14$, and $19$ points as the dynamic density grows while $R_l$ improves only marginally, so predicting risk and using risk are distinct mechanisms.

Removing the risk field altogether reduces the method to a dual-stream variant of a purely cost-supervised one-stage planner. With no spatially structured target, the representation is shaped only by the scalar trajectory cost: success falls to $36\%$ in 60 Dynamic while $v_a$ rises to $5.02\,\mathrm{m/s}$, the signature of a policy flying straight and fast into unanticipated hazards.

Removing the explicit stream causes the largest drop, to $26\%$. The risk head must then regress a metric angular quantity from optical-axis depth, whose peripheral distortion is what the range map removes, so predictions are poorly calibrated at the FoV boundary where crossing obstacles first appear, and the coupling propagates these errors into the trajectory cost, inflating $R_l$ to $1.23$. FRM therefore makes the risk field learnable, the privileged target makes it correct, and the coupling makes it consequential.

\begin{table}[t]
    \setlength{\tabcolsep}{5pt}
    \footnotesize
    \centering
    \caption{Ablation study on the explicit range-map stream, the spatio-temporal risk field, and the arrival-time risk coupling}
    \begin{threeparttable}
    {
    \begin{tabular}{m{1.0cm}<{\centering}m{1.95cm}<{\centering}m{0.9cm}<{\centering}m{1.2cm}<{\centering}m{0.9cm}<{\centering}}
       \toprule[1.2pt]
       \textbf{Scene} & \textbf{Method} & \textbf{$\eta(\%)$} & \textbf{$v_a(m/s)$} & \textbf{$R_l$} \\
       \midrule[0.8pt]

    \multirow{4.5}{*}{\rotatebox[origin=c]{90}{\shortstack{40 Static \&\\20 Dynamic}}}
     & Proposed & \best{89} & 5.46 & 1.06 \\
     & \emph{w/o} $J_{\text{rf}}$ & \secondbest{79} & \secondbest{5.58} & \secondbest{1.05} \\
     & \emph{w/o} ST-RF & 72 & \best{5.63} & \best{1.04} \\
     & \emph{w/o} FRM & 66 & 5.19 & 1.10 \\
     \midrule

    \multirow{4.5}{*}{\rotatebox[origin=c]{90}{\shortstack{20 Static \&\\40 Dynamic}}}
     & Proposed & \best{78} & 4.73 & 1.12 \\
     & \emph{w/o} $J_{\text{rf}}$ & \secondbest{64} & \secondbest{4.86} & \secondbest{1.10} \\
     & \emph{w/o} ST-RF & 54 & \best{4.92} & \best{1.08} \\
     & \emph{w/o} FRM & 42 & 4.38 & 1.18 \\
     \midrule

    \multirow{4.5}{*}{\rotatebox[origin=c]{90}{\shortstack{60\\Dynamic}}}
     & Proposed & \best{67} & 4.58 & 1.14 \\
     & \emph{w/o} $J_{\text{rf}}$ & \secondbest{48} & \secondbest{4.69} & \secondbest{1.12} \\
     & \emph{w/o} ST-RF & 36 & \best{5.02} & \best{1.09} \\
     & \emph{w/o} FRM & 26 & 4.21 & 1.23 \\
     \bottomrule[1.2pt]
    \end{tabular}
    }
    \end{threeparttable}
    \label{tbl:ablation_study}
\end{table}

\begin{figure*}[!t]
    \centering
    \includegraphics[width=0.85\linewidth]{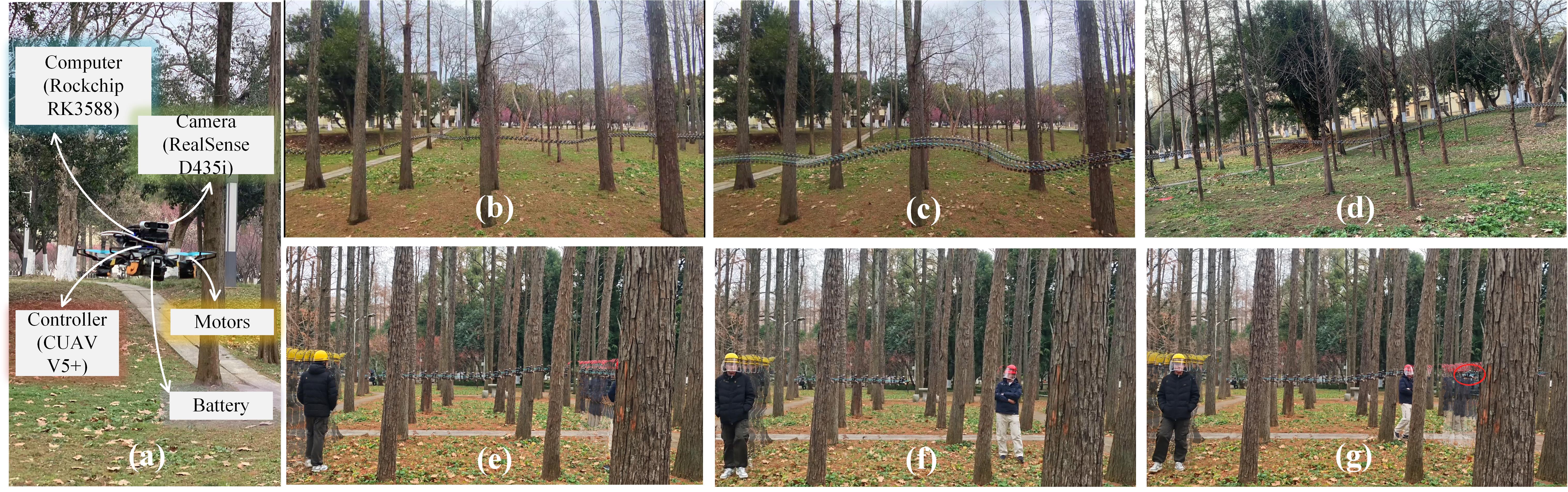}
    \caption{\textbf{Real-world experiments.} (a) Our quadrotor platform: a Rockchip RK3588 onboard computer, an Intel RealSense D435i, a CUAV V5+ flight controller, the motors, and the battery. (b)--(d) Static flights through a dense forest, with (d) crossing a sloped section of irregular trunk spacing. (e)--(g) Dynamic flights in the same forest with two walking pedestrians; the red ellipse in (g) marks the quadrotor. Every sequence is composited from a fixed viewpoint, so the repeated silhouettes trace the executed trajectory.}
    \label{fig:real}
\end{figure*}

\begin{table}[!t]
    \centering
    \caption{Latency of each component in the proposed system}
    \begin{tabular}{l c c c}
       \toprule[1.2pt]
       \textbf{System Modules} & \textbf{\shortstack{RTX 3090 Ti\\(ms)}} & \textbf{\shortstack{RK3588\\(ms)}} & \textbf{\shortstack{Orin NX\\(ms)}} \\
       \midrule[0.8pt]
       Depth Processing        & 1.02 & 1.71 & 1.19 \\
       Frustum Range Map       & 0.58 & 1.09 & 0.79 \\
       Data Preparation        & 0.55 & 1.26 & 0.82 \\
       Network Inference       & 4.81 & 9.02 & 5.94 \\
       Proposal Selection      & 0.41 & 0.91 & 0.57 \\
       \midrule
       \textbf{Total Latency}  & \textbf{7.37} & \textbf{13.99} & \textbf{9.31} \\
       \bottomrule[1.2pt]
    \end{tabular}
    \label{tbl:latency}
\end{table}

\subsection{Real-World Experiments}
The platform is shown in Fig.~\ref{fig:real}(a): a Rockchip RK3588 onboard computer, an Intel RealSense D435i, and a CUAV V5+ flight controller, with a Jetson Orin NX as a second computing option. The network is quantized to INT8 on the RK3588 NPU and compiled to a TensorRT engine with FP16 kernels on the Orin NX, and at test time the planner relies solely on onboard sensing at $50~\mathrm{Hz}$, with the privileged target, the ESDF, and all actor states absent and the risk head retained for~\eqref{eq:score_mod}. In Table~\ref{tbl:latency}, network inference dominates, while the range-map branch adds at most $1.09\,\mathrm{ms}$ because it is a min-pooling reduction rather than a learned module, and both onboard totals stay inside the $20\,\mathrm{ms}$ budget implied by the replanning rate.

\subsubsection{Static scenes}
The trained policy is deployed directly in a previously unseen forest, Fig.~\ref{fig:real}(b)--(d). Without explicit mapping, the quadrotor threads collision-free corridors between the trunks, and the composited trajectories stay smooth rather than piecewise reactive, including on the sloped section of (d). The conservative min-pooled range map is the main contributor, since it filters raw sensor artifacts into the abstraction used in simulation and narrows the reality gap without domain randomization.

\subsubsection{Dynamic scenes}
We further evaluate dynamic avoidance in the same forest with two walking pedestrians, Fig.~\ref{fig:real}(e)--(g). From the dual-stream observation the quadrotor infers pedestrian motion and steers toward directions whose predicted risk stays low over the horizon: as a pedestrian moves into the planned corridor, avoidance begins while the separation is still large, and the vehicle passes behind the pedestrian rather than braking in place. The self-predicted field is observable onboard, making these decisions inspectable at runtime.

\section{Conclusion}
\label{sec:conclusion}
In this paper, we present RiskFly, a one-stage learning-based planner for agile UAV navigation in dynamic and cluttered environments, which predicts risk in the same space in which it acts.
A dual-stream observation pairs a depth sequence with a frustum-aligned inverted spherical range-map sequence, so that perception, risk prediction, and end-state proposals share one angular grid.
On this grid, an auxiliary head regresses a spatio-temporal risk field supervised by a privileged closest-point-of-approach target, and the self-predicted field is queried differentiably at the arrival times of the instantiated trajectory, so supervision and planning gradients act on one shared field while the policy remains expert-free and map-free.
Extensive experiments in simulation and real flights demonstrate higher success rates at comparable latency.
Future work will replace the constant-velocity extrapolation with a learned forecast and extend the risk field to omnidirectional sensing.

\bibliographystyle{IEEEtran}
\bibliography{references}

\end{document}